\documentclass{IOS-Book-Article}

\usepackage{hyperref}
\usepackage{mathptmx}
\usepackage{listings}
\usepackage{caption}
\usepackage{subcaption}
\usepackage{graphicx}
\usepackage{amsmath,amsthm,amssymb}
\usepackage{todonotes}
\newtheorem{definition}{Definition}
\graphicspath{ {./img/} }

\def\hb{\hbox to 10.7 cm{}}

\begin{document}

\pagestyle{headings}
\def\thepage{}
\long\def\ignore#1{}

\begin{frontmatter}              

\title{The NAI Suite -- Drafting and Reasoning over Legal Texts}
\runningtitle{Drafting and Reasoning over Legal Texts}

\author[A]{\fnms{Tomer} \snm{Libal}}
and
\author[B]{\fnms{Alexander} \snm{Steen}}

\runningauthor{Libal and Steen}
\address[A]{American University of Paris}
\address[B]{University of Luxembourg}

\begin{abstract}
  A prototype for automated reasoning over legal texts, called NAI, is presented.
  As an input, NAI accepts formalized logical representations of such legal texts
  that can be created and curated using an integrated annotation interface.
  The prototype supports automated reasoning over the given text representation
  and multiple quality assurance procedures.
  The pragmatics of the NAI suite as well its feasibility in practical applications
  is studied on a fragment of the \textit{Smoking Prohibition (Children in Motor Vehicles) (Scotland) Act 2016}
  of the Scottish Parliament.
\end{abstract}

\end{frontmatter}

\section{Introduction}

Computer systems are playing a substantial role in assisting people in a wide range of tasks,
including search in large data and decision-making; and their employment
is progressively becoming vital in an increasing number of fields.
One of these fields is \textit{legal reasoning}:
New court cases and legislations are accumulated every day. In addition, international
organizations like the European Union are constantly aiming at combining and integrating
separate legal systems~\cite{burley1993europe}.
In contrast to this situation, the automation of legal reasoning is still underdeveloped
albeit being a growing field of research.
In recent years automatic procedures, e.g. for courtroom
management\footnote{
  See \url{http://softpert.com/legal/court-management/winjuris}.
} and legal language processing/management~\cite{boella2016eunomos},
expert systems based on cases or rules~\cite{zeleznikow1995split}, 
and normative compliance tools\footnote{
See \url{https://cst.cnpd.lu/portal} for GDPR compliance checking.
} have been introduced.
At the same time, approaches for automatic reasoning over sets of norms have been developed,
such as in the courtroom~\cite{aucher19}, for the HIPAA and GLBA privacy laws~\cite{deyoung2010experiences},
for business compliance~\cite{HashmiG18} and GDPR compliance~\cite{palmirani2018modelling}.

One of the main reasons for the relatively restricted number of applications of automated reasoning to
the legal domain is the lack of editing tools which can be used by non-logicians. Indeed, the applications
mentioned above are mainly based on the work of logicians. In order to have a wider use of legal reasoning,
other professionals, such as lawyers and jurists, should be able to use the tools.

A second reason is the lack of tools and methodologies for asserting the correctness of the logical representations
of the legal texts. Among existing results, one can find a methodology for building legal ontologies \cite{mockus2017legal}
and more concretely to our approach, one for validating formal representations of legal texts \cite{bartolini2018interdisciplinary}.

Lastly, the scarcity of legal reasoning software prevents the utilization of such formalizations, even if proven correct.
One can mention here the engines for defeasible logics \cite{governatori2013regorous}, Higher-order logics \cite{J46}
and Deontic logics with contrary-to-duty obligations \cite{LibalP19}.

In this paper we describe the new normative reasoning framework NAI,
which addresses these problems.
NAI is a web application and is readily available at \url{https://nai.uni.lu}.
NAI is also open-source, its source code is freely available at GitHub\footnote{
  See \url{https://github.com/normativeai}.
} under GPL-3.0 license.

NAI features an annotation-based editor which abstracts over the underlining logical language.
It also contains an easily accessible functionality for ensuring that the formalization is consistent and that the
formalized sentences are independent from each other. NAI also supports a methodology for proving
the correctness of formalizations via execution of behavioral tests. Lastly, it provides an interface for the creation of queries
and for checking their validity.

The architecture of NAI is modular, which allows using different logics and reasoning engines. It also provides an API,
which can be used by other tools in order to reason over the formalized legislation.

The contributions of the paper are: A technical description of a new tool for legal formalization and reasoning
which utilizes an innovative annotations interface.
A user guide for the use of this tool. A description of a novel methodology for checking the correctness of
formalizations, based on ideas from software engineering and the execution of tests.
Finally, we apply the tool to a fragment of a legal text - we apply the methodology
to formalize an article and we use the tool for reasoning automatically
over different use cases.

In \S\ref{sec:prelim} and \S\ref{sec:nai}, the logical foundations and features, respectively, of
the NAI suite are presented. Subsequently, \S\ref{sec:example} presents a prototypical case study
on the application of NAI on a concrete legal text. Finally, \S\ref{sec:conc} concludes and
sketches further work.

\section{Preliminaries \label{sec:prelim}}
The logical formalism underlying the NAI framework is based on a universal fragment first-order variant of
the deontic logic \textbf{DL}*~\cite{LibalP19}, denoted \textbf{DL}*$_1$. Its syntax is given by
\begin{definition}[Syntax of \textbf{DL}*$_1$]
Let $V$, $P$ and $F$ be disjoint sets of symbols for variables, predicate symbols (of some arity)
and function symbols (of some arity), respectively.
\textbf{DL}*$_1$ formulas $\phi, \psi$ are given by:
\begin{equation*}\begin{split}
  \phi,\psi & ::= p(t_1, \ldots, t_n)
         \; | \; \neg \phi
         \; | \; \phi \land \psi
         \; | \; \phi \lor \psi
         \; | \; \phi \Rightarrow \psi \\
        &\quad | \; \mathsf{Id}\,\phi
         \; | \; \mathsf{Ob}\,\phi
         \; | \; \mathsf{Pm}\,\phi
         \; | \; \mathsf{Fb}\,\phi
         \; | \; \phi \Rightarrow_{\mathsf{Ob}} \psi
         \; | \; \phi \Rightarrow_{\mathsf{Pm}} \psi
         \; | \; \phi \Rightarrow_{\mathsf{Fb}} \psi
\end{split}\end{equation*}
where $p \in P$ is a predicate symbol of arity $n \geq 0$ and the $t_i$, $1 \leq i \leq n$, are terms.
Terms are freely generated by the function symbols from $F$ and variables from $V$.
\mbox{} \hfill $\lrcorner$
\end{definition}
\textbf{DL}*$_1$ extends Standard Deontic Logic (SDL) with the normative concepts  of ideal and
contrary-to-duty obligations, and contains predicate symbols, the standard logical connectives,
and the normative operators of obligation ($\mathsf{Ob}$), permission
($\mathsf{Pm}$), prohibition ($\mathsf{Fb}$), their conditional counter-parts, and ideality ($\mathsf{Id}$).
Free variables are implicitly universally quantified at top-level.

This logic is expressive enough to capture many interesting normative structures.
For details on its expressivity and its semantics, we refer to previous work~\cite{LibalP19}.

\section{The NAI Suite \label{sec:nai}}

\begin{figure}[t]
\centering
\includegraphics[width=.4\linewidth,interpolate]{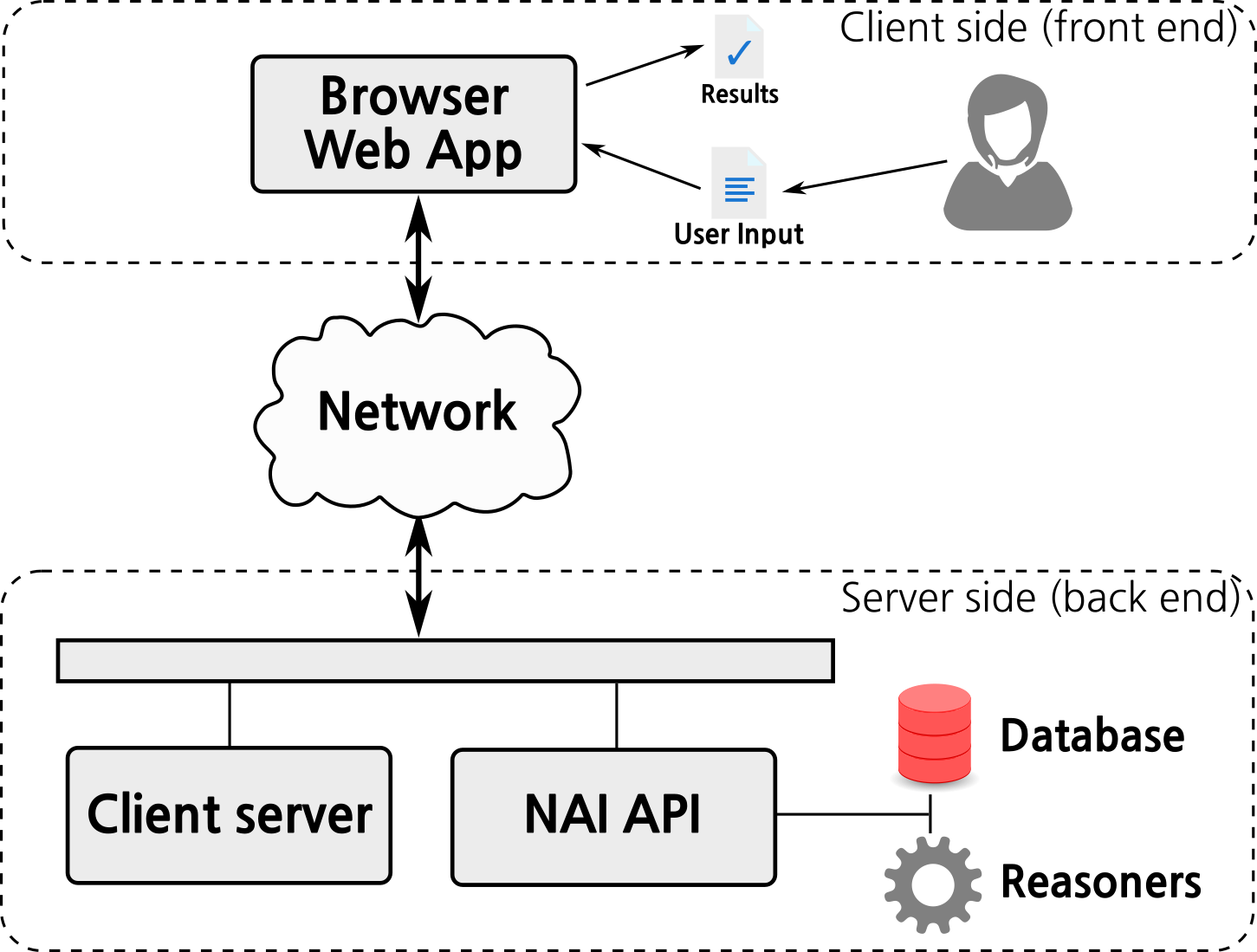}
\caption{Software-as-a-service architecture of the NAI reasoning framework. The front end
software runs in the user's browser and connects to the remote site, and its different
services, via a well-defined API through the network. Data flow is indicated by
arrows. \label{fig:architecture}}
\end{figure}

The NAI suite integrates novel theorem proving technology
into a usable graphical user interface (GUI) for the computer-assisted formalization
of legal texts and applying automated normative reasoning procedures on these artifacts.
In particular, NAI includes
\begin{enumerate}
  \item a legislation editor that graphically supports the formalization of legal texts,
  \item means of assessing the quality of entered formalizations, e.g., by automatically
        conducting consistency checks and assessing logical independence,
  \item ready-to-use theorem prover technology for evaluating user-specified
        queries wrt. a given formalization, and
  \item the possibility to share and collaborate, and to experiment with different formalizations
        and underlying logics.
\end{enumerate}
NAI is realized using a web-based Software-as-a-service architecture, cf. Fig.~\ref{fig:architecture}.
It comprises a GUI that is implemented as a Javascript browser application,
and a NodeJS application on the back-end side which connects to theorem provers, data storage services
and relevant middleware. Using this architectural layout, no further software
is required from the user perspective for using NAI and its reasoning procedures, as all
necessary software is made available on the back end and the
computationally heavy tasks are executed on the remote servers only. The results of the
different reasoning procedures are sent back to the GUI and displayed to the user.
The major components of NAI are described in more detail in the following.


\subsection{The Reasoning Module}
The NAI suite supports formalizing legal texts and applying various logical operations on them.
These operations include consistency checks (non-derivability of falsum), logical independence
analysis as well as the creation of user queries that can automatically be assessed for
(non-)validity. After formalization, the formal representation of the legal text is
stored in a general and expressive machine-readable format in NAI. This format aims at
generalizing from concrete logical formalisms that are used for evaluating the logical
properties of the legal document's formal representation.

There exist many different logical formalisms that have been discussed for capturing
normative reasoning and extensions of it. Since the discussion of such formalisms is still ongoing,
and the choice of the concrete logic underlying the reasoning process strongly influences the results
of all procedures, NAI uses a two-step procedure to employ automated reasoning tools.
NAI stores only the general format, as mentioned above, as result of the formalization process.
Once a user then chooses a certain logic for conducting the logical analysis, NAI will automatically
translate the general format into the specific logic resp. the concrete input format of the
employed automated reasoning system.
Currently, NAI supports only the \textbf{DL}*$_1$ logic from \S\ref{sec:prelim}; however, the architecture
of NAI is designed in such a way that further formalisms can easily be supported.
Possible extensions are described in \S\ref{sec:conc}.

The choice in favor of \textbf{DL}*$_1$ is primarily motivated by the fact that it can be effectively
automated using a shallow semantical embedding into normal (bi-)modal logic~\cite{LibalP19}. This
enables the use of readily available reasoning systems for such logics; in contrast, there are
few to none automated reasoning systems available for normative logics (with the exception of \cite{governatori2013regorous}).
In NAI, we use
the MleanCoP prover~\cite{Otten14} for first-order multi-modal logics as it is currently one
of the most effective systems and it returns proof certificates which can be independently
assessed for correctness~\cite{Otten12}. It is also possible to use various different tools
for automated reasoning in parallel (where applicable). This is of increasing importance once
multiple different logical formalisms are supported.

\subsection{The Annotation Editor} \label{sec:annotation}

The annotation editor of NAI is one of its central components. Using the editor,
users can create formalizations of legal documents that can subsequently used
for formal legal reasoning. The general functionality of the editor is described in
the following. A more detailed exemplary application on a concrete legal document is presented
in \S\ref{sec:example}.

One of the main ideas of the NAI editor is to hide the underlying logical details and technical
reasoning input and outputs from the user. We consider this essential, as the primary target audience
of the NAI suite are not necessarily logicians and it could greatly decrease the usability of
the tool if a solid knowledge about formal logic was required.
This is realized by letting the user annotate legal texts and queries graphically and
by allowing the user to access the different reasoning functionalities by simply clicking buttons
that are integrated into the GUI.
Note that the user can still inspect the logical formulae that result from the annotation process
and also input these formulae directly. However, this feature is considered advanced and not the
primary approach put forward by NAI.

The formalization proceeds as follows:
The user selects some text from the legal document and annotates it, either as a
term or as a composite (complex) statement. In the first case, a name for that term is
computed automatically, but it can also be chosen freely. Different terms are displayed as
different colors in the text.
In the latter case, the user needs to choose among the different possibilities
(which roughly correspond to logical connectives) and the containing text can be annotated
recursively.
Composite statements are displayed as a box around the text. An example of an annotation
result is displayed in Fig.~\ref{fig:annot}


The editor also features direct access to the consistency check and logical independence check procedures (as buttons).
When such a button is clicked, the current state of the formalization will be translated and sent
to the back-end provers, which determine whether it is consistent resp. logically independent.

User queries are also created using such an editor. In addition to the steps sketched above,
users may declare a text passage as \emph{goal} using a dedicated annotation button, whose contents
are again annotated as usual.
If the query is executed, the back-end provers will try to prove (or refute) that the goal logically follows
from the remaining annotations and the underlying legislation.

\subsection{The Abstract Programming Interface (API)}

All the reasoning features of NAI can also be accessed by third-party applications.
The NAI suite exposes a RESTful (Representational state transfer) API which allows (external) applications
to run consistency checks, checks for independence as well as queries and use the result
for further processing.
The exposure of NAI's REST API is particularly interesting for external legal applications that want to
make use of the already formalized legal documents hosted by NAI.
A simple example of such an application is a tax counseling web site which advises its visitors
using legal reasoning over a formalization of the relevant tax law done in the NAI suite.

\section{Case Study: Scottish Smoking Regulation}
\label{sec:example}

In this section we are going to demonstrate how the NAI suite can be used on a legal text.
The text we will use is the "Smoking Prohibition (Children in Motor Vehicles) (Scotland) Act 2016" \footnote{\url{https://www.legislation.gov.uk/asp/2016/3/contents}}.
We have chosen this text as it makes a perfect candidate for legal reasoning, being short and relatively self contained.
It has also featured in previous research \cite{wyner2017annotation}.

This legislation contains 19 articles which go from describing the conditions of committing the offence
to how a fine can be given and contested.

In this example, we will focus on article 1 only. A more comprehensive formalization which includes sentences of the second
part as well, is available online \footnote{Please visit \url{nai.uni.lu} and log in with the credentials: smoking@nai.lu / nai}. \\

\textbf{Article 1: Offence of smoking in a motor vehicle with children}
\begin{enumerate}
\item It is an offence for an adult to smoke in a private motor vehicle when: (a) there is a child in the vehicle, and (b) the vehicle is in a public place.
\item Subsection (1) does not apply to a private motor vehicle that is designed or adapted for use as living accommodation and which, at the time the smoking occurs, is parked and is being used as living accommodation.
\item A person who commits an offence under subsection (1) is liable on summary conviction to a fine not exceeding level 3 on the standard scale.
\end{enumerate}

In order to be able to apply automated reasoning to this text, we first need to formalize our understanding of its meaning. In
other words, we need to formalize a legal interpretation of the text.

There are various interpretations possible even for this, relatively simple, text. For the purpose of this example,
we interpret the article as prohibiting adults to smoke in a private motor vehicle in case:
(1) there is a child in the vehicle,
(2) the vehicle is in public space and
(3) the vehicle is not adapted or designed to be used, and at the same time is being used, as living accommodation.

Violating this prohibition, the adult is liable to a fine via a summary conviction.

\subsubsection*{The methodology}

The formalization process is essentially translating an informal natural language text into a formal logical formula or code.
As mentioned before, this step is essential for being able to apply automated reasoning techniques.

We can choose various formulae in the logic \textbf{DL}*$_1$ which seem to describe the text above. Each of these formulae
differs in the cases it holds and in the consequences which can be derived from it.

A correct formalization means that the right formula is chosen. How can we pick this formula? In \cite{bartolini2018interdisciplinary},
Bartolini \& al. define a methodology for the validation of the formal representation of legal texts by a backward translation
to a human-readable text. The text is then being validated by legal experts. Mockus and Palmirani \cite{mockus2017legal} define
a method for the iterative refinement of ontologies, which is inspired by a previous work by Peroni \cite{peroni2016simplified}.
Peroni's work adapts approaches from the agile methodology in software engineering. The above approaches still depend on
humans for validation. In this section we describe a new methodology which is based on
Behavior Driven Development (BDD) \footnote{\url{https://www.agilealliance.org/glossary/bdd/}}. The "behaviors"
defined by this methodology are validated by machines, similarly to those in software engineering.

BDD is a methodology for developing code
which follows several steps. First, the client writes down informal texts which define
what the code is supposed to do. These texts, called user stories or behaviors, should be as comprehensive as possible.
In the next step, the user stories are being translated into many test programs. These tests are supposed to pass when
the relevant code is implemented and satisfies the tests. The tests are normally being generated semi-automatically from the user stories.
In the last step, the programmer is left with the relatively easy task of writing just enough code to pass all the tests.

This methodology enjoys very high success in software engineering and we believe that it can be adapted to legal text formalization
as follows.

The lawyer writes down different scenarios which should be true (or false), given her interpretation of the legal text.
The lawyer then annotates these scenarios in order to translate them into test formulae. In the last step, a person needs
to annotate the legal text in a way such that all the test formulae will be validated. It should be noted that the person
in the last step must not have a full legal understanding of the text and that in principle, this last step can even be executed by
a machine, which tries different formalization possibilities until all test formulae are satisfied.

We need therefore to start with a comprehensive list of scenarios and their outcomes based on our legal interpretation. It should be
noted that such scenarios are normally based on many articles or even on the whole text. In our example, we will derive them from
article 1 only.

Here we describe just a few of these scenarios. The reader is referred to the live example in the application for more cases.

\begin{figure}
\centering
\begin{subfigure}{.5\textwidth}
  \centering
  \includegraphics[width=0.94\linewidth,interpolate]{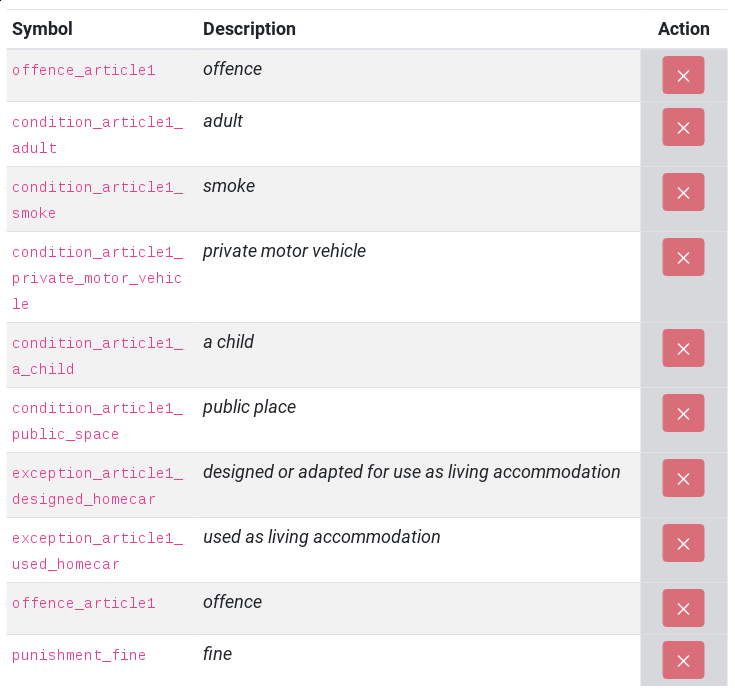}
  \caption{Vocabulary}
  \label{fig:voc}
\end{subfigure}%
\begin{subfigure}{.5\textwidth}
  \centering
  \includegraphics[width=0.94\linewidth,interpolate]{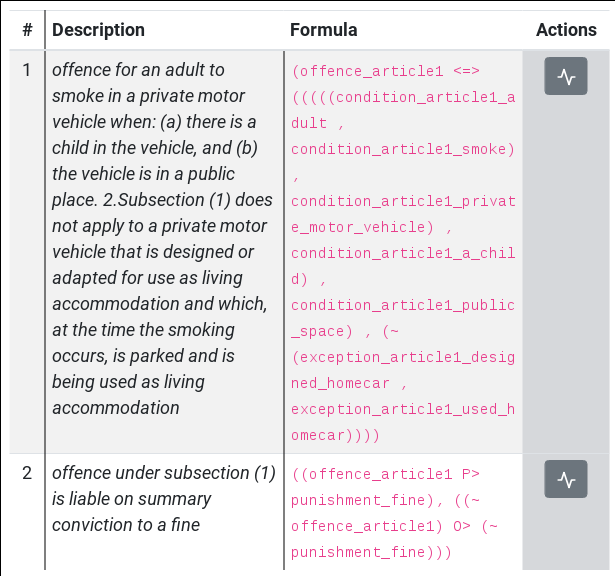}
  \caption{\textbf{DL}*$_1$ formulae}
  \label{fig:form}
\end{subfigure}
  \caption{Smoking legislation article 1}
\end{figure}

The first step in the methodology is to create the vocabulary used in the formalization. As mentioned in Section \ref{sec:annotation},
this is being done by using the \emph{term} annotation on the text. The annotated terms can then be seen on the
"Vocabulary" tab of the NAI suite. Figure \ref{fig:voc} summarizes those for Article 1.

The test queries can now be created based on this vocabulary. The task of the lawyer is to consider different terms from the vocabulary
and decide what is the expected outcome of them.

\textbf{Scenario 1.} An adult was smoking in a car which has a child in it and is not in public space. We expect the adult not to be liable to a fine.

\textbf{Scenario 2.} An adult was smoking in a car which has a child in it, is in public space and was not designed as living accommodation.
We expect the adult to be liable to a fine.

The lawyer now uses the queries tab in the NAI suite in order to enter these two scenarios. In order to differentiate
the test queries from case queries (queries written in order to solve a specific case), the test queries names are prefixed
with "Test ".

\begin{figure}
\centering
\begin{subfigure}{.5\textwidth}
  \centering
  \includegraphics[width=0.94\linewidth,interpolate]{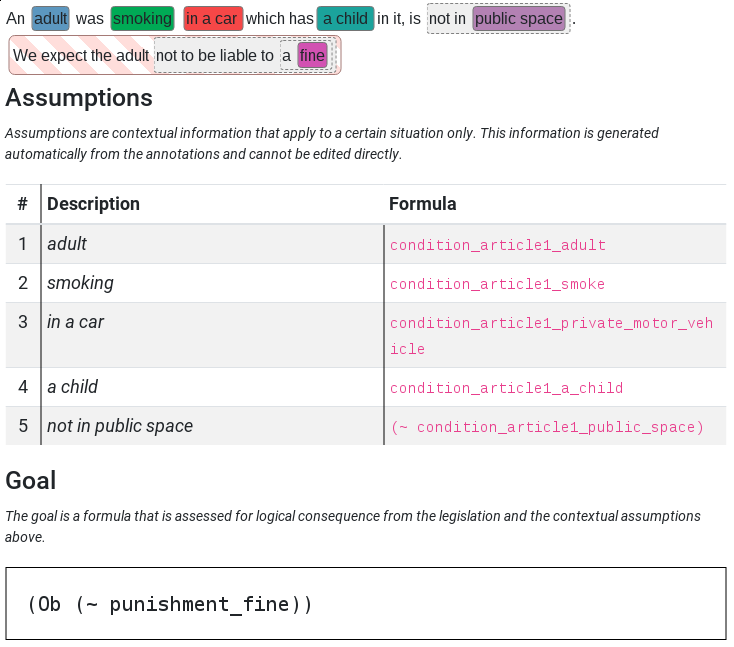}
  \caption{Scenario 1.}
  \label{fig:test1}
\end{subfigure}%
\begin{subfigure}{.5\textwidth}
  \centering
  \includegraphics[width=0.94\linewidth,interpolate]{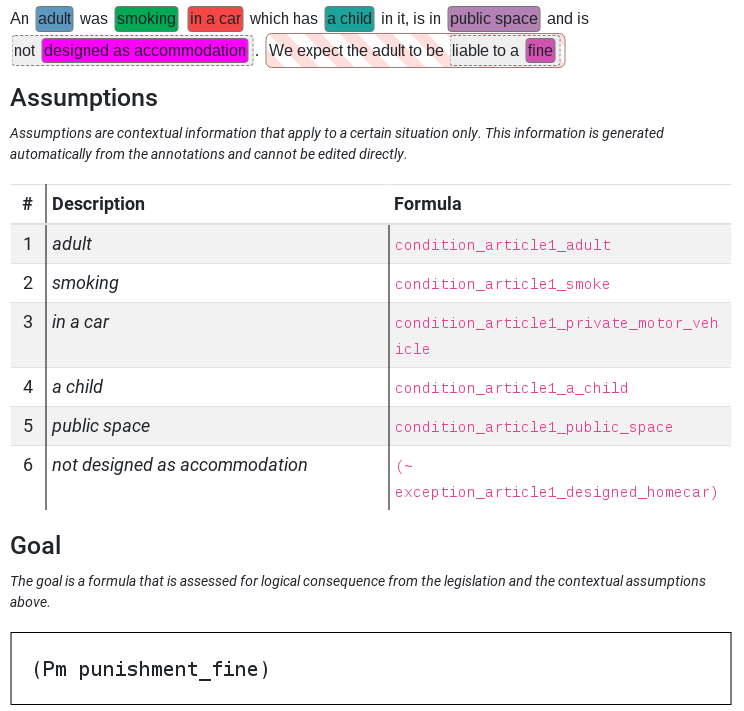}
  \caption{Scenario 2.}
  \label{fig:test2}
\end{subfigure}
  \caption{Annotations and \textbf{DL}*$_1$ formulae.}
\end{figure}

We can now annotate the two scenarios.
We proceed first by annotating the conditions with the terms from the vocabulary. The user needs to select those from a drop down list.
The expectation is then annotated as a goal. Within the goal, we annotate our expectation that the person is liable to a fine by using
the \emph{Permission} connective over the \emph{punishment\_fine} term. The two annotated scenarios, as well as their formalization,
can be seen in figures
\ref{fig:test1} and \ref{fig:test2}. When executing these queries, they naturally may fail. When annotating the legal text
in the next phase, we must make sure that all the queries are now being validated.

\begin{figure}
  \centering
  \includegraphics[width=0.45\linewidth,interpolate]{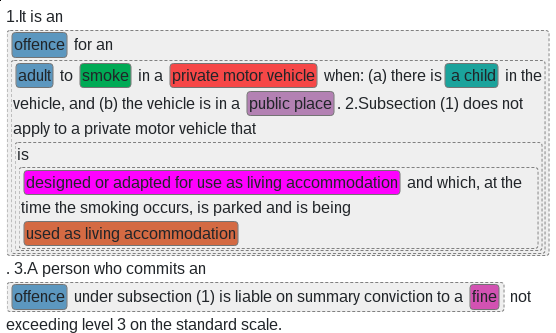}
  \caption{Smoking legilsation article 1: annotation}
  \label{fig:annot}
\end{figure}

We can now proceed with the last step - the annotation of Article 1. After some trial and error, we have ended up
with the annotation in Figure \ref{fig:annot}. This annotation passes all of our test queries and we therefore
conclude that it is a faithful formalization of our interpretation of Article 1. The \textbf{DL}*$_1$ formulae are shown in Figure \ref{fig:form}.

It should be mentioned that on each step, we are advised to check the consistency of our annotations as well as those of the queries.
The reasoning engine can find automatically inconsistencies in our annotations, which can lead to wrong results. In addition, it is
recommended to check, on the "Formalization" tab, that each \textbf{DL}*$_1$ formula is independent. Dependent formulae are normally a sign
of an incorrect formalization.

\subsubsection*{Case queries}

Once we are confident that our formalization is faithful to our interpretation, we can trust it to resolve legal questions
with regard to specific cases. Writing case queries is identical to the writing of test queries. As an example,
consider the following case.

\textbf{Case 1.} A client got a fine while driving his home car while smoking. His teen daughter was sitting next to him.
Is there a case to appeal this decision?

Here we want to check if there was an obligation in the law not to give our client the fine.
In case it is true, an appeal should be successful.
When we annotate the case above, we get that a conclusion cannot be drawn (the query is counter-satisfiable).
The reason for that is because some of the conditions are not used. Since there might be two different values
to these conditions which result in two different conclusions, the reasoner cannot determine if the query holds.
In this case, we can find in the "Vocabulary" tab one further condition - the car should be in public space - and one further exception
- the car should also be used as a home car, and not only be designed as one. We therefore ask the client to share more information
about the case.

\textbf{Case 2.} The client adds further that he was indeed driving in public space. The home car though, was not used
as a home car at the time. The client has removed the home facilities and is using the car for transportation of goods.

The addition of the new annotations gives us the answer that the policeman was indeed permitted to give the fine.
The client could enjoy the exception of subsection (b), but he failed to use the car for accommodation. It seems
better not to appeal the fine.

\section{Conclusion and Future Work}
\label{sec:conc}

In this paper we have described the NAI suite, introduced a new methodology and showed how it can be used for generating
correct formalizations.
We have demonstrated on a relatively simple case, how these formalizations can be used in order to help lawyers in their work.

Furthermore, the use cases show how the tool can be used by non-experts as well. No legal knowledge is required
in order to write down and reason over specific cases. In fact, one can argue that the legal expert is essential only when writing down
an extensive list of test queries. Given those, a formalization can be generated automatically by enumerating
all possibilities and executing the tests over them.

The ability of non-experts to use the tool can be exploited further by using NAI's API. Given a formalization of a legal interpretation
which is done by a professional, web applications can be created which allow the user to input cases and obtain legal consequences
based on it. Since the reasoning engines produce a proof of correctness, this automatic legal counsel can be vouched for by the professional.
Such applications can help customers get a legal counsel regarding matters such as air passenger rights and citizenship, or even
be used in order to automatize some legal procedures.

The tools presented in this paper are prototypes. Further work is required on both the tools and their
supporting theories in order to make the formalization of legal texts easier and more intuitive.
Among those improvements, the most notable ones relate to the supporting theory and to the usability of the user interface.
We mention several such improvements here.

Currently, the NAI suite supports an expressive deontic first-order language. This language is rich enough to describe many
scenarios which appear in legal texts. Nevertheless, more work is required in order to capture all such scenarios.
Among those features with the highest priority, we list support for exceptions, temporal sentences and arithmetic.
In this paper,
we overcame the fact that subsection 1(b) is an exception to subsection 1(a) by explicitly mentioning the values of the
conditions of the exception. This solution is not optimal since it requires the setting of values to these properties in all
tests and cases. Possible support for these features already exists in the form of tools such as non-monotonic reasoners \cite{kifer2005nonmonotonic},
temporal provers \cite{suda2012pltl} and SMT solvers \cite{bouton2009verit}.

On the level of usability, the tool currently does not give any information as to why a query is counter-satisfiable. The user needs
to look on the vocabulary in order to determine possible reasons. Integrating a model finder, such as Nitpick \cite{blanchette2010nitpick},
will help "debugging" formalizations.

NAI's graphical user interface (GUI) aims at being intuitive and easy to use and tries to hide the underline complexities
of the logics involved. A continuously updated list of new features can be found on the GUI's development website \footnote{\url{https://github.com/normativeai/frontend/issues}} .

\bibliographystyle{plain}
\bibliography{paper}

\end{document}